\documentclass{article}

\usepackage{arxiv}

\usepackage[utf8]{inputenc} % allow utf-8 input

\usepackage{newtxtext,newtxmath}
\usepackage[T1]{fontenc}   % better font encoding
\usepackage{microtype}     % nicer kerning & justification

\usepackage{hyperref}       % hyperlinks
\usepackage{url}            % simple URL typesetting
\usepackage{booktabs}       % professional-quality tables
\usepackage{amsfonts}       % blackboard math symbols
\usepackage{nicefrac}       % compact symbols for 1/2, etc.
\usepackage{microtype}      % microtypography
\usepackage{lipsum}		% Can be removed after putting your text content
\usepackage{graphicx}
\usepackage[authoryear,round]{natbib}
\usepackage{doi}
\usepackage{tabularx}

\let\cite\citep

\graphicspath{ {graphics/} }

\newenvironment{supplement}{
  \setcounter{figure}{0}\setcounter{table}{0}\setcounter{equation}{0}%
}{}

\title{Balancing Exploration and Exploitation in Low-N Protein Engineering}

\renewcommand{\headeright}{}   % remove top-right text

\makeatletter
\renewcommand{\maketitle}{
  \begin{center}
    {\LARGE \bfseries \@title \par}
    \vskip 1em
    {\@author \par}
    \vskip 0.5em
  \end{center}
}
\makeatother

\title{Low-N Protein Activity Optimization with FolDE}

\makeatletter
\renewcommand{\and}{,\penalty-1\hskip 0.5em plus 0.3em minus 0.3em}
\makeatother

\author{
Jacob B. Roberts$^{1,2,3}$\and
Catherine R. Ji$^{4}$\and
Isaac Donnell$^{1,3,5}$\and
Thomas D. Young$^{1,3}$\and
Allison N. Pearson$^{1,3,6}$\and
Graham A. Hudson$^{1,3,7}$\and
Leah S. Keiser$^{1,3,8}$\and
Mia Wesselkamper$^{1,9}$\and
Peter H. Winegar$^{1,3,7}$\and
Janik Ludwig$^{1,3,10}$\and
Sarah H. Klass$^{1,3,8}$\and
Isha V. Sheth$^{1,11}$\and
Ezechinyere C. Ukabiala$^{1,2,3}$\and
Maria C. T. Astolfi$^{1,2,3}$\and
Benjamin Eysenbach$^{4,\#}$\and
Jay D. Keasling$^{1,2,3,5,7,8,12,\#}$}

\date{}

\begin{document}
\maketitle
\thispagestyle{empty}
% \setlength{\droptitle}{-2em} % pulls title block up
% \vspace{-1.2em}              % tighten space after authors

\begin{footnotesize}
$^{1}$ Joint BioEnergy Institute, Emeryville, CA, USA \\
$^{2}$ Department of Bioengineering, UCSF / UCB \\
$^{3}$ Biological Systems and Engineering, Lawrence Berkeley National Laboratory, Berkeley, CA, USA \\
$^{4}$ Princeton University \\
$^{5}$ Chemistry Department, UCB \\
$^{6}$ Department of Plant and Microbial Biology, UCB \\
$^{7}$ QB3 Institute, University of California, Berkeley, CA, USA \\
$^{8}$ Department of Chemical and Biomolecular Engineering, UCB \\
$^{9}$ Bioengineering Department, UCB \\
$^{10}$ Faculty of Biology, Ludwig-Maximilians-Universität München \\
$^{11}$ Chemical Biology Department, UCB \\
$^{12}$ The Novo Nordisk Foundation Center for Biosustainability, Technical University Denmark, Kemitorvet, Kongens Lyngby, Denmark \\
$^{\#}$~Corresponding authors: Jay D. Keasling (\texttt{jdkeasling@lbl.gov}); Benjamin Eysenbach (\texttt{eysenbach@princeton.edu})\end{footnotesize}

\vspace{10pt}

% \begin{document}
% \maketitle

\begin{abstract}
	Proteins are traditionally optimized through the costly construction and measurement of many mutants. Active Learning-assisted Directed Evolution (ALDE) alleviates that cost by predicting the best improvements and iteratively testing mutants to inform predictions. However, existing ALDE methods face a critical limitation: selecting the highest-predicted mutants in each round yields homogeneous training data insufficient for accurate prediction models in subsequent rounds. Here we present FolDE, an ALDE method designed to maximize end-of-campaign success. In simulations across 20 protein targets, FolDE discovers 23\% more top 10\% mutants than the best baseline method (p=0.005) and is 55\% more likely to find top 1\% mutants. FolDE achieves this primarily through naturalness-based warm-starting, which augments limited activity measurements with protein language model outputs to improve activity prediction. We also introduce a constant-liar batch selector, which improves batch diversity; this is important in multi-mutation campaigns but had limited effect in our benchmarks. The complete workflow is freely available as open-source software, making efficient protein optimization accessible to any laboratory.

\end{abstract}

% keywords can be removed
\keywords{Low-N Protein Engineering \and ALDE \and Reinforcement Learning}

\section{Introduction}
Proteins power processes in medicine, manufacturing, and agriculture. The relationship between a protein's sequence and its activity defines a complex, often rugged landscape \citep{Wu2016-yc,Meger2024-zh,Husain2020-di}. Protein language models (PLMs) – neural networks trained on databases of naturally occurring protein sequences – have emerged as powerful tools for navigating these landscapes \citep{Jiang2025-pb,Yang2025-jj,Meier2021-wt,Alley2019-fd}. PLMs learn to predict how likely a given amino acid sequence is to occur in nature. They provide two key capabilities. First, the relative probability a PLM assigns to a mutated sequence, which we term its naturalness, correlates with protein activity \citep{Hie2022-kw,Meier2021-wt}. Second, PLMs can convert variable-length sequences into fixed-length embeddings, which can be used to extrapolate between sequences with known activity \citep{Alley2019-fd}.

Directed evolution improves protein activity through sequential rounds of mutation and screening or selection \citep{Selles-Vidal2023-dx}. Traditional campaigns require thousands to millions of mutants, but many targets lack high-throughput screens, limiting researchers to dozens of mutants. Active learning for directed evolution (ALDE) methods make optimization feasible under these constraints by using computational models to select batches of mutants between rounds \citep{Yang2025-jj}. State-of-the-art ALDE methods achieve impressive results: EVOLVEpro successfully improved five targets with as few as 16 mutants per round \citep{Jiang2025-pb}. The methods are varied but follow key patterns. Most begin with randomly selected mutants in the first round \citep{Bao2022-te}; others begin with the top ranked naturalness mutants \citep{Singh2025-jg}. In later rounds, they predict activities for a pool of prospective mutants, often using PLM embeddings, then select the highest ranked (top-N selection). While successful, current approaches leave unresolved tradeoffs. There is no consensus on the impact of starting a campaign with naturalness-based mutants, and there is no clearly successful mechanism for diversifying selected mutants while continuing to search for top performers.

\begin{figure}
	\centering
	% \fbox{\rule[-.5cm]{4cm}{4cm} \rule[-.5cm]{4cm}{0cm}}
    \includegraphics[width=\linewidth]{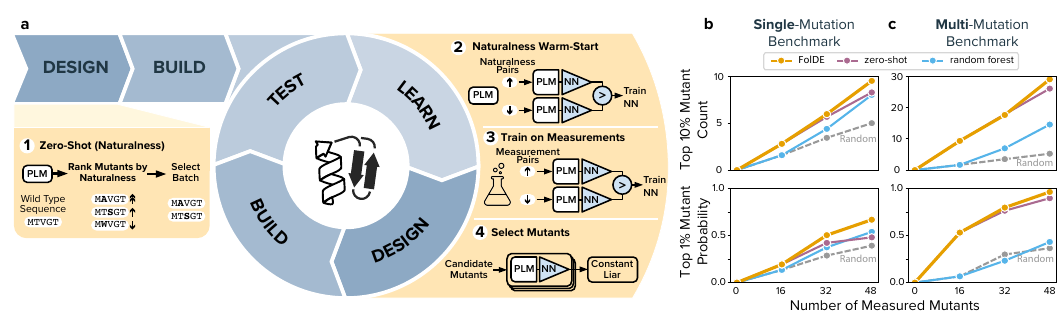}
	\caption{\textbf{The FolDE Workflow} (a) schematic of the FolDE workflow, starting with the zero-shot prediction before data has been collected, followed by few-shot prediction and Bayesian batch building, in a design-build-test-learn cycle. (b) Performance on the single-mutation and (c) multi-mutation benchmark for FolDE vs three baselines: random selection, zero-shot naturalness-based selection, and a random forest with embeddings (representing EVOLVEpro, \citet{Jiang2025-pb}). Metrics shown are the cumulative top 10\% mutants discovered (top) and probability of finding a top 1\% mutant (bottom). }
	\label{fig:fig1}
\end{figure}

We consider protein optimization as an iterative, closed loop process in which decisions in each round determine the data available in later rounds. Through this lens, we identify two failure modes of current methods that occur in such small-budget protein optimization campaigns. First, there is tension between round-1 and round-2 model performance. Starting a campaign with naturalness-based zero-shot selection in round-1 provides much better mutants than random selection (3.8× more top 10\% mutants, 3.6× higher chance of finding a top 1\% mutant, Fig. \ref{fig:fig1}b,c), but this early exploitation comes at a cost. The resulting mutants are clustered in high naturalness regions of sequence space, providing insufficient diversity to train conventional activity prediction models (like random forest) in round-2. As a result, starting a campaign with naturalness-based selection yields better mutants in round-1, but no better than random mutants in round-2. Second, in later rounds, the highest ranked mutants selected are often slight variants of top performing mutants from previous rounds, providing little new information about the activity landscape. These failure modes are manifestations of a classic exploration-exploitation tradeoff \citep{Berger-Tal2014-ea} in an iterative, data-limited problem.

We present the FolDE (Foldy's Directed Evolution) protein optimization method, which avoids some pitfalls of over-exploration or -exploitation. FolDE includes two new policies: incorporation of PLM outputs in activity prediction with naturalness warm-starting and diversity-aware batch selection (Fig. \ref{fig:fig1}a). In simulation, we compared FolDE to a random forest ALDE baseline (representing EVOLVEpro): FolDE discovers about 23\% more top 10\% mutants in rounds 1-3 (p=0.005), and is 55\% more likely to find a top 1\% mutant. FolDE is released as open-source software with a user interface, making sophisticated protein optimization accessible to any lab.

\section{Main Text}

\subsection{Experimental Design}

To develop and evaluate protein optimization workflows, we created an iterative simulation benchmark using pre-collected data from ProteinGym \citep{Notin2023-rd}. We partitioned ProteinGym datasets into a training set of 9 proteins for workflow development and a test set comprising 17 single-mutation and 3 multi-mutation datasets (Table S1). In each simulation, workflows iteratively select batches of 16 mutants across three rounds (48 total mutants), mimicking real low-throughput protein engineering campaigns. FolDE was developed and optimized using the training datasets, and then compared to baselines on the single- and multi-mutation sets.

Real protein optimization campaigns routinely explore sequence space six or more mutations away from the wild-type sequence, with some campaigns venturing as far as 29 mutations \citep{Patsch2024-dg,Cobb2013-ac}. While comprehensive datasets at these depths remain unavailable, our multi-mutation benchmarks provide a better approximation of real campaigns than single-mutation datasets. The multi-mutation datasets include double mutants and contain hundreds of thousands of mutants, compared to only thousands in single-mutation datasets. This larger combinatorial space better captures the challenge of building useful batches from many similar-looking candidate mutants, a key difficulty in real campaigns exploring three or more mutations.

We measure success by two metrics that directly reflect protein optimization goals: the cumulative number of top 10\% mutants discovered and the probability of finding at least one top 1\% mutant within three rounds. These metrics capture both the overall quality of the selected batch (how many good mutants did we find) and success at the critical task of discovering exceptional mutants, making them more relevant to practical protein optimization than correlation-based metrics like Spearman $\rho$. We benchmark the FolDE workflow against three baselines: random selection (representing traditional directed evolution), zero-shot naturalness-based selection (using PLMs without learning between rounds), and random selection in round-1 followed by random forest with ESM2-15b embeddings (representing the EVOLVEpro workflow, \citet{Jiang2025-pb}).

\subsection{Iterative Protein Optimization Workflow}

Our aim is to build a system that optimizes protein activity through iterative rounds of selecting and testing batches of mutants. The key challenge, as we show below, is selecting proteins that both have high activity and provide valuable information for selecting even better proteins in future rounds. FolDE, the method we developed, comprises five key components: (1) zero-shot naturalness-based selection in round-1, (2) neural network with ranking loss, (3) naturalness warm-start, (4) ensemble predictions, and (5) constant-liar batch selection with $\alpha=6$ in round-2.

In round-1, we use naturalness-based zero-shot selection (Fig. \ref{fig:fig1}a, Step 1). We compute wild-type marginal likelihood (“naturalness”) with ESM-family protein language models \citep{Lin2022-xg,Hie2024-ic}, and select for screening the mutants with the top ranked naturalness.

In subsequent rounds, we train an activity prediction model composed of a PLM to embed protein sequences followed by an activity-predicting neural network (Fig. \ref{fig:fig1}a, Step 2). This architecture, a PLM for embedding followed by a top-layer to predict activity, has a long history in protein activity prediction. The top-layer is often a random forest model, which has shown strong performance in activity regression, outperforming neural networks \citep{Jiang2025-pb}. Recently, \citep{Brookes2024-fz} demonstrated that neural networks trained with ranking loss outperform neural networks trained with regression for protein activity prediction. We find that a neural network top-layer trained with ranking loss finds slightly more top 10\% mutants than random forest, and many more than a neural network trained with regressive loss (Fig. S1). We train multiple networks to obtain an ensemble of predictions, which enables computation of both mean and covariance of predicted ranks \citep{Gawlikowski2023-mw}.

The neural network undergoes two training phases (Fig. \ref{fig:fig1}a, Step 3). First, weights are warm-started using the same naturalness predictions that were used for zero-shot selection, training on all possible single mutants. Recent work has shown that incorporation of naturalness as a prior improves activity prediction \citep{Meier2021-wt,Krause2022-fm,Hawkins-Hooker2024-il,Zhou2024-jv,Lafita2024-ko}, and we adapt this insight through a warm-start approach that distills PLM priors into a compact neural network top layer. Second, the network is fine-tuned on accumulated activity measurements. The ability to train the neural net on two different classes of data is facilitated by the use of ranking loss, which is invariant to the scale or units of the training data.

Finally, we predict activities for all candidate mutant sequences and select a batch of sequences using constant-liar (Fig. \ref{fig:fig1}a, Step 4). While previous work has encouraged exploration during selection through Bayesian methods like upper confidence bound (UCB) \citep{Yang2025-jj,Hu2023-uy,Hie2022-zs}, TuRBO \citep{Cheng2022-ad}, and Thompson sampling \citep{Yang2025-jj}, none to our knowledge employ batch-aware acquisition strategies. Constant-liar is a Bayesian optimization approach that approximately solves parallel expected improvement by iteratively building diverse batches. After selecting the highest-scoring mutant, it assumes that mutant performs poorly – this pessimistic assumption propagates through the prediction ensemble's covariance structure, downweighting predictions for similar mutants in subsequent selections. The parameter $\alpha$ controls exploitation intensity, with high values minimizing the lie's impact and recovering top-N selection. For the three-round benchmark simulations presented in this work, FolDE uses $\alpha=6.0$ in round-2 only, providing a balance between the exploitation needed to find high-performing mutants and the exploration needed to build robust models for future rounds. In extended simulations examining batch diversity over more rounds, constant-liar continues to be applied in round-3 and beyond. We investigate the effect of varying $\alpha$ on batch diversity and model performance in the section below.

\subsection{Naturalness Warm-Start Alleviates Round-1 / Round-2 Tension}

\begin{figure}
	\centering
	% \fbox{\rule[-.5cm]{4cm}{4cm} \rule[-.5cm]{4cm}{0cm}}
    \includegraphics[width=\linewidth]{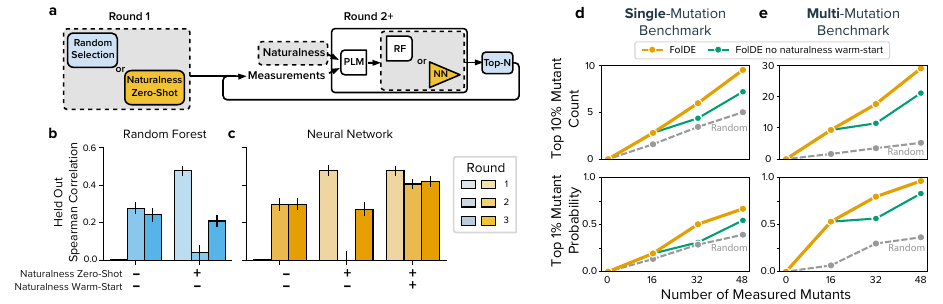}
	\caption{\textbf{The Apparent Tension Between Round-1 and Round-2 Explore and Exploit}  (a) The simple workflow under study, notably excluding some FolDE features like ensembling and constant-liar. We study the interplay between three features: choice of top-layer (random forest or a neural network), round-1 selection approach (random selection or naturalness zero-shot selection), and the inclusion of naturalness warm-start when training. (b) training benchmark experiment results: the Spearman correlation of the trained model on a held-out set of mutants for three rounds of simulation for a random forest top-layer and (c) for a neural network top-layer. (d) Performance of the FolDE workflow without the warm-start feature enabled, measured on the single-mutation benchmark and (e) multi-mutation benchmark.
}
	\label{fig:fig2}
\end{figure}

To understand how mutant selection in one round affects model performance in subsequent rounds, we evaluated a baseline ALDE method, EVOLVEpro, with and without the use of naturalness-based mutant selection in round-1. We observed that naturalness correlates with activity (Spearman $\rho \approx 0.48$, Fig. \ref{fig:fig2}b, round 1) and using naturalness in round-1 improves the selected mutants. But using naturalness in round-1 affects the training data available in round-2. We observed that models trained with naturalness-selected mutants have a collapse in prediction quality relative to those trained with randomly selected mutants (Spearman $\rho$: 0.04 vs. 0.27, Fig. \ref{fig:fig2}b round 2). This suggests a tension between optimizing immediate returns and building effective models for future rounds. The FolDE architecture, without warm-start, suffers the same round-2 collapse (Fig. \ref{fig:fig2}b, columns 1\&2). We hypothesized that this performance degradation stems from data bias: when all training examples are high-naturalness mutants, the model has insufficient information about the broader mutant landscape.

To address this data bias problem, we developed a naturalness warm-starting procedure. Naturalness serves two purposes in our workflow: first, we continue to select round-1 mutants based on their naturalness scores, providing strong initial performance. Second, we now pretrain the activity prediction neural network to recapitulate naturalness scores for all possible single mutants before fine-tuning it on the limited activity measurements from round-1. We were inspired by recent work that has shown improvements in activity prediction by incorporating naturalness. For example, \cite{Groth2024-gr} incorporate naturalness as a prior in Gaussian-process based activity prediction, and \cite{Hawkins-Hooker2024-il,Krause2022-fm} always select mutants based on naturalness, but they fine-tune their PLM with activity data. Our distillation approach differs in that we compress the large PLM's predictions into a compact neural network, which provides computational advantages for rapid ensembling and the extensive simulations presented here. While distillation preserves the naturalness rankings, it was not obvious that this pretraining step would address the round-2 prediction collapse caused by biased round-1 data. However, by training on all single mutants rather than just the high-naturalness mutants selected in round-1, the procedure exposes the model to the full spectrum of mutant quality. 

We find that naturalness warm-start preserves the model's predictive accuracy across all three rounds: whereas models without warm-start suffer a collapse in round-2 correlation (Spearman $\rho$ dropping from 0.48 to 0.04), warm-started models maintain strong correlation throughout ($\rho \approx 0.48$ in rounds 1-3; Fig. \ref{fig:fig2}c). In the combined test benchmark, this improved prediction quality translates to substantially better mutant discovery, with warm-started FolDE discovering 32\% more top 10\% mutants and achieving a 23\% higher probability of finding a top 1\% mutant compared to FolDE without warm-start (Fig. \ref{fig:fig2}d,e).

\subsection{Homogeneous Later Round Batches}

\begin{figure}
	\centering
	% \fbox{\rule[-.5cm]{4cm}{4cm} \rule[-.5cm]{4cm}{0cm}}
    \includegraphics[width=\linewidth]{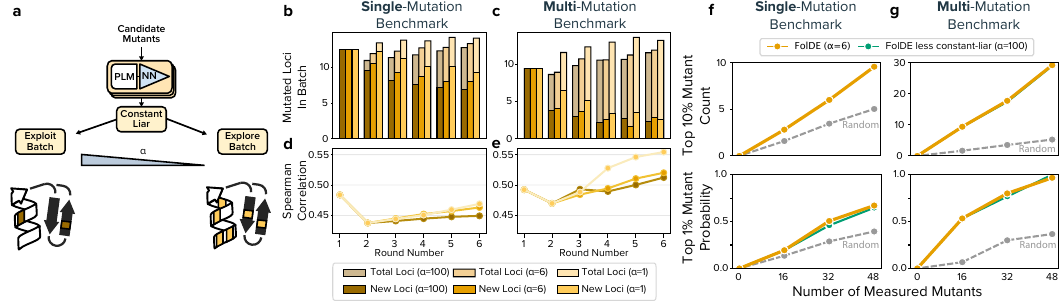}
	\caption{\textbf{Constant-Liar Improves Batch Diversity} (a) Schematic of the constant-liar algorithm. After selecting a high-performing mutant from the pool, the algorithm pessimistically assumes that mutant performs poorly (the "lie"), propagating this assumption through the prediction ensemble's covariance structure. The alpha parameter controls the balance between exploitation and exploration, with lower values creating more diverse batches. (b) Batch diversity for the single-mutation and (c) multi-mutation datasets with constant-liar applied for six rounds. Bars indicate the number of unique loci sampled per batch, with darker colors showing newly explored loci. (d) Model predictions are more accurate with more aggressive constant-liar on the single-mutation and (e) multi-mutation benchmarks. (f) Medium constant-liar ($\alpha=6$) applied in round-2 slightly improves the probability of finding a top 1\% mutant in the single mutation benchmark. (g) Medium constant-liar ($\alpha=6$) in round-2  has little effect on the probability of finding a top 1\% mutant in the multi-mutation benchmark.}
	\label{fig:fig3}
\end{figure}

We observed that top-N selection produces homogeneous batches that are concentrated on previously successful locations in the protein (loci). For example, in round-2 of the multi-mutation benchmark, for a batch of 16 mutants with two mutations each, only 4 out of 32 mutations targeted new loci (Fig. \ref{fig:fig3}c). Diverse batches offer several advantages: they provide richer training data for future rounds, enable stacking of beneficial mutations from different loci, provide robustness against measurement errors, and reduce the risk of converging on local optima.

Prior work has applied Bayesian methods like upper confidence bound (UCB) \citep{Yang2025-jj,Hu2023-uy,Hie2022-zs}, TuRBO \citep{Cheng2022-ad}, and Thompson sampling \citep{Yang2025-jj} to balance exploration and exploitation. However, these methods select mutants based on individual merit without considering within-batch similarity – UCB simply selects the 16 highest-scoring mutants regardless of their structural redundancy. We found UCB has no effect on batch diversity (Fig. S2). We therefore introduce constant-liar for batch selection, described above, which explicitly promotes diversity through pessimistic assumptions about selected mutants. Constant-liar is parameterized by $\alpha$ where high values lead to pure exploitation and low values encourage more exploration.

Constant-liar successfully improves batch diversity. In extended simulations of the single- and multi-mutation benchmarks, the number of new loci in the round-2 batches increased from 9.5 and 3.8 with $\alpha=100$ to 12.2 and 7.3 with $\alpha=1.0$ (Fig. \ref{fig:fig3}b-c). In these extended simulations, enhanced diversity translates to modest improvements in model predictions on held out variants, with more aggressive constant-liar (lower $\alpha$) yielding better Spearman correlations (Fig. \ref{fig:fig3}d,e).

To select an appropriate $\alpha$ value, we evaluated performance on the training benchmark using both cumulative top 10\% hits and the probability of finding a top 1\% mutant. We found that $\alpha$ values between 3 and 6 slightly improve the probability of finding a top 1\% mutant while slightly decreasing the number of 10\% hits discovered (Fig. S3). Based on this analysis, FolDE uses $\alpha=6$ in round-2, reverting to pure exploitation ($\alpha=100$) in round-3. On the single- and multi-mutation test benchmarks, constant-liar shows minimal impact on the primary metrics (Fig. \ref{fig:fig3}f,g), though these metrics may underestimate the benefits in real campaigns which are exploring exponentially larger sequence spaces.

\subsection{Benchmark Results}
We benchmarked FolDE against three baselines on 17 single-mutation and 3 multi-mutation ProteinGym datasets that were not used in model building (Sup. Table 1). The baseline methods are: (1) random selection (representing traditional directed evolution) \citep{Yang2025-jj}, (2) zero-shot naturalness-based selection (using PLMs without learning between rounds), and (3) random selection in round-1 folled by the ESM2-15b PLM for embeddings and a random forest ("random forest", representing the EVOLVEpro workflow) \citep{Jiang2025-pb}. The simulations were run for 3 rounds at 16 mutants per round, and evaluated on the cumulative number of top 10\% mutants discovered and probability of finding a top 1\% mutant. Across all 20 test set proteins (Fig. S4), FolDE discovered in median 23\% more top 10\% mutants than random forest (p=0.005, one-sided Wilcoxon test, range: 0.62×-2.32×), and also was in median 55\% more likely to find a top 1\% mutant (p=0.037, range: 0.25×-3.0×). FolDE also outperforms zero-shot selection, with improvements growing in later rounds, discovering 6\% more 10\% mutants (p=0.009, range: 0.83×-13.6×) and 15\% higher probability of finding a top 1\% mutant (p=0.003). Notably, zero-shot fails to find any top 1\% mutants in 5 of 20 targets, whereas FolDE achieves nonzero probability for all targets.

On the single mutation benchmark, FolDE increases the number of top 10\% mutants discovered by 13\% over random forest (p=0.032, range: 0.62×-2.14×), with improvements in 13 out of 17 targets (Fig. S4). FolDE also improves the probability of discovering at least one top 1\% mutant within three rounds by 50\%, with improvements in 11 out of 17 targets.

The multi-mutation benchmark datasets are much larger, and are more similar to a real campaign. This benchmark includes the pairwise mutation landscape of three multimutant datasets from ProteinGym (Sup. Table 1). Relative to random forest, FolDE found a median of 96\% more top 10\% mutants (range: 1.7×-2.3×), and increased the median probability of finding a top 1\% mutant by 2.25× (range: 2.0×-2.5×) (Fig. S4). FolDE did not improve much on the zero-shot baseline in the multi-mutation benchmark - both methods selected very high quality batches and had a high probability of finding a top 1\% mutant in the first three rounds (Fig. \ref{fig:fig1}c).

\begin{figure}
	\centering
	% \fbox{\rule[-.5cm]{4cm}{4cm} \rule[-.5cm]{4cm}{0cm}}
    \includegraphics[width=300pt]{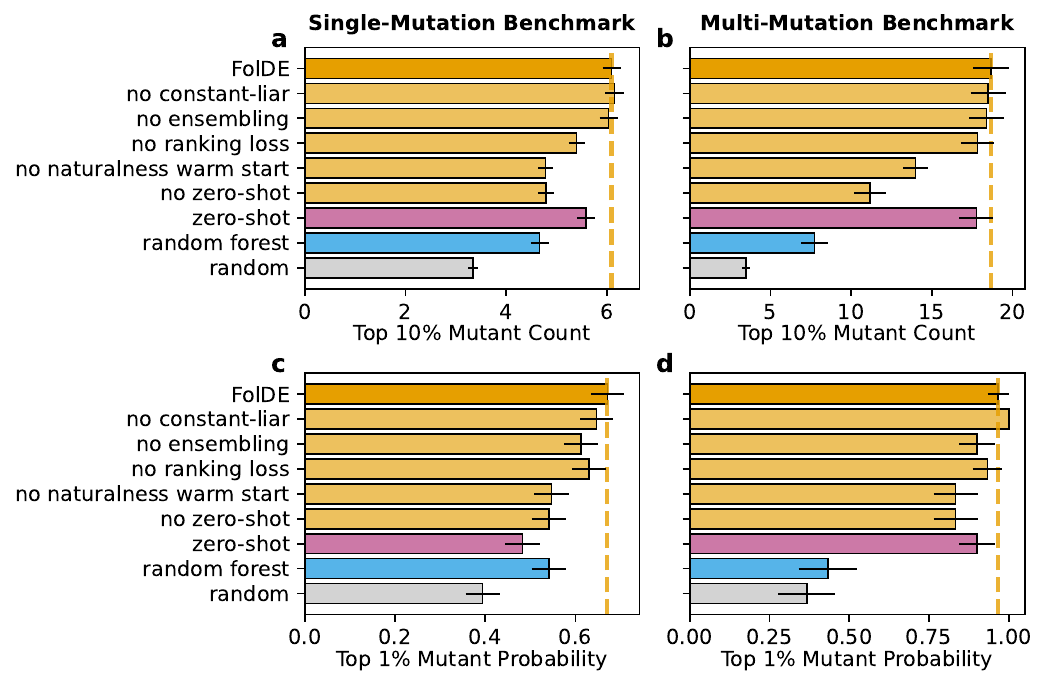}
	\caption{\textbf{FolDE Ablation}   (a) The relative contribution of the major workflow components were evaluated by the number of top 10\% mutants discovered after a 3 round campaign on both the single-mutation benchmark and (b) the multi-mutation benchmark. (c, d) 3 round probability of finding a top 1\% mutation.}
	\label{fig:fig4}
\end{figure}

We evaluated which components of the FolDE workflow contributed to the improved performance, as measured on both the single- and multi-mutation test benchmarks (Fig. \ref{fig:fig4}). The components that most contribute to performance are zero-shot prediction in round-1, naturalness warm-start, and the use of a ranking loss. Constant-liar, with the settings used in the simulation, did not have a large effect on these top line metrics.

\subsection{Discussion}
The protein optimization task is a data-limited and iterative optimization process. We identified two examples of ways that optimization in the short term leads to model degradation in later rounds. First, zero-shot selected mutants have higher activity, but when taken as training data for top-layer style models the models perform poorly. Second, especially in the multi-mutation optimization space, top-N selection yields homogeneous mutant batches which are poor training data for later rounds. These pitfalls support the use of iterative benchmarks when evaluating activity predictors for low-N protein optimization.

Scientific foundation models, including protein language models, are emerging as powerful tools for scientific discovery, but realizing their potential requires an iterative lens. Our findings demonstrate that low-N refinement of foundation model predictions through active learning can dramatically improve experimental outcomes. These same constraints and opportunities may apply to other foundation models for regulatory elements, genomes, and whole-cell models.

The efficiency gains from FolDE enable deeper exploration of mutation space, moving beyond three mutations to engineer more radical functional changes with the same experimental budget. By making these methods accessible through the open-source software, we lower the barrier for labs without specialized machine learning expertise to perform efficient protein optimization. With a 55\% higher chance of finding a top 1\% mutant in three rounds, FolDE moves protein optimization from a resource-intensive endeavor toward a routine optimization that any researcher can perform in a matter of months.

\section{Methods}

\subsection{The FolDE Architecture}

\textbf{Zero-Shot Mode} FolDE (Foldy's Directed Evolution) operates through iterative design-build-test-learn cycles to optimize protein activity with minimal experimental measurements. The workflow has two distinct operational modes: zero-shot selection for the first round when no activity data exists, and few-shot learning for subsequent rounds. For simulations, in each round FolDE selects 16 mutants for experimental characterization, and uses these measurements to improve predictions in future rounds.

Before any activity measurements are collected, FolDE uses protein language model naturalness scores to select promising mutants. We calculate naturalness as the log-likelihood ratio of mutant to wild-type amino acids. Let $x$ be the wild type sequence and $x'$ be the mutant sequence with mutations at locations $T$. The protein language model takes in an unmasked sequence $x$ and parameters $\theta$ and returns the probability distribution over amino acids at all locations $P(y|x,\theta)$ \citep{Gordon2024-vh,Hawkins-Hooker2024-il}. The naturalness is as follows:

\begin{align}
\text{Naturalness}(x', x) = \sum_{i \in T} \log P(x'_i | x, \theta) - \log P(x_i | x, \theta)
\label{eq:naturalness}
\end{align}

This formulation for naturalness requires a single pass through the PLM. One trait of this formulation is that separate mutations are regarded independently, and their effects are summed in log space. This formulation was found to be just as accurate are more compute-intensive formulations which allow for non-linear (epistatic) mutation interactions, with the exception of the "modulo" masking strategy which is slightly more accurate with hundreds of times greater compute cost \cite{Hawkins-Hooker2024-il}. In effect, this formulation scores mutations highly where the mutant looks good and the wild type sequence looks bad, and this calculation requires no biological or experimental information – all insights come from the unsupervised PLM training.

For the first round we rank all single mutants by naturalness and select the top 16 mutants. In practice but not in simulation, we selected at most 3 mutations per locus to encourage diversity.

\textbf{Few-Shot Mode} In round-2 and beyond, after collecting activity measurements, FolDE transitions to few-shot learning using an ensemble of neural networks. The workflow has three steps: naturalness warm-start \& activity training, predicting the activity of candidate mutants, and constant-liar batch building (Fig. \ref{fig:fig1}a, 2-4).

\textbf{Embedding} All amino acid sequences are preprocessed into an embedding vector using a PLM. To construct the embedding vector we do a forward pass through the PLM and mean-pool the final hidden layer \citep{Vieira2025-ul}. We found that all ESM2 and ESMC architectures with 150M parameters or larger had similar performance (Fig. S5). For FolDE we use ESMC-300M for embedding, which is a 50x reduction in compute cost relative to ESM2-15b.

\textbf{Model Initialization} We initialize an ensemble of five MLPs. Each ensemble member is a multi-layer perceptron mapping 960-dimensional protein embeddings to scalar activity predictions:
\begin{itemize}
    \item Input: 960-dimensional ESMC-300M embeddings
    \item Hidden layers: 960 → 100 → 50 → 1
    \item Batch normalization after each hidden layer
    \item ReLU activations and dropout (p=0.2)
    \item No bias in the final linear layer
    \item Bradley-Terry ranking loss
    \item Random initial weights.
\end{itemize}

\textbf{Learning Protocol} Models are optimized using Adam (learning rate $3e^{-4}$, weight decay $1e^{-5}$) with automatic mixed precision for computational efficiency. During warm-start we train for a maximum of 50 epochs with early stopping based on validation loss (patience of 20 epochs, validation every 5 epochs). During activity training we train for a maximum of 200 epochs with early stopping based on validation loss (patience of 40 epochs, validation every 10 epochs). The Bradley-Terry loss computes probabilities for all directed pairs within a batch, weighted by binary cross-entropy between predicted and true rankings.

In both training steps, data is used to construct all directed pairs, and those pairs are split 80-20 into train and validation sets. The split is done so the training set does not trivially solve any pairs in the validation set. For example, if the training set contains both pairs (A,B) and (B,C) where A > B and B > C, then the validation set will not include the pair (A, C). This is done with a breadth-first search. With this split, all data points can contribute to training and validation loss can still give a reflection of the training quality.

\textbf{Naturalness Warm-Start} Before training on activity data, we pretrain each ensemble member for 50 epochs to approximate the naturalness landscape. We found, on the training benchmark, that any number of warm-start epochs more than 10 was sufficient to capture the warm-start improvement (Fig. S6). The same naturalness is used in warm-start as in zero-shot. The formula is given in Equation \eqref{eq:naturalness} (above). This pretraining uses embeddings of all single mutants as inputs and their corresponding naturalness scores as targets, providing the model with information about broadly tolerated substitutions that might be absent from biased first-round selections. 

\textbf{Activity Training} We fine-tune the warm-started models on measured activities using a Bradley-Terry ranking loss, which learns from pairwise comparisons. This approach is well suited to protein optimization for several reasons. First, identifying high performing mutants requires accurate ranking of mutants rather than prediction of activity; for this reason, BT loss improves protein activity prediction in the context of global epistasis \citep{Brookes2024-fz}. Second, ranking loss is invariant to the scale of the training data, enabling our two-phase training procedure where naturalness scores (warm-start) and activity measurements have different units and ranges. For validation, we construct a directed acyclic graph of training pairs and select validation pairs that cannot be trivially inferred through transitivity, ensuring meaningful performance assessment.

\textbf{Ensemble Configuration} We train 5 independent models with different random seeds, computing consensus predictions as the mean across models. To compensate for the fact that Bradley-Terry loss (BT) is translation invariant, we de-mean within-ensemble predictions before computing the consensus. This ensemble approach provides uncertainty estimates and improves prediction robustness.

\textbf{Activity Prediction} Each round expands the search space by considering single mutants of any previously measured high-performing mutant. Round-2 considers mutations of wild-type and round-1 hits; round-3 adds mutations of round-2 hits, allowing progressive exploration while maintaining the ability to backtrack. We use the ensemble of models to create an ensemble of activity predictions for each candidate sequence.

\textbf{Constant-Liar} The constant-liar (CL) algorithm is a method to increase batch diversity during parallel evaluation while maximizing a target (e.g. activity) on Gaussian Processes \citep{Ginsbourger2008-bv,Ginsbourger2010-or,Wu2016-qg}. Given a set of candidates $x_i$ and labels $y_i$, CL greedily chooses the UCB-maximizing candidate $(x_i, y_i)$ as the first element in the batch. Then, CL updates the covariance matrix and mean over the remaining candidates with an imagined observation of the greedy candidate.

Consider a set of  $N$ mutants $M = \{m_1, \dots, m_N\}$. Let $\mathbf{\Sigma}$ be the covariance matrix of predicted activity levels of $M$ over models trained with different seed initializations and ensembling. Let $\mathbf{v}$ be the covariance between $M \setminus m_i$ and $m_i$ and $\mathbf{\Sigma}_{M \setminus m_i}$ be the covariance matrix over the remaining $M \setminus m_i$ mutants. 

The matrix $\mathbf{\Sigma}$ takes the following form, where $i$ is used as shorthand for $m_i$. Let $y_M = \{y_1, \dots, y_N\}$ denote the estimated activity levels of the $N$ mutants:

\begin{align}
\begin{pmatrix}
    \mathbf{\Sigma}_{M \setminus i} & \mathbf{v} \\
    \mathbf{v}^{T} & \sigma^2_{i}.
\end{pmatrix}
\end{align}

with the covariance matrix posterior

\begin{align}
\mathbf{\Sigma}' = \mathbf{\Sigma}_{M \setminus i} - \mathbf{v} \mathbf{v}^{T} / \sigma^2_{i}.
\end{align}

The covariance matrix update is independent of the value of the lie.

The mean value posterior depends on the value of the lie $y^{\text{lie}}_i$:

\begin{align}
y'_{M \setminus i} = y_{M \setminus i} + \mathbf{v} (y^{\text{lie}}_i - y_i) / \sigma^2_{i}.
\end{align}

We use a pessimistic lie $\min{y_i}$ within the set of candidate labels. In the implemented version of CL, this update corresponds to the Bayesian-optimal posterior of a Gaussian Process given the imagined observation $y_i^{\text{lie}}$.

Because the imagined values are pessimistic, CL is incentivized to explore candidates distinct from the top-ranked candidate. Future work should explore whether more theoretically motivated methods like parallel knowledge gradient \citep{Wu2016-qg} lead to improved performance in the search for high-activity mutants. CL-type and, more generally, Kriging Believer algorithms have been shown to be simple and robust in practical experimental-design settings \citep{Ginsbourger2008-bv,Ginsbourger2010-or,Wu2016-qg,Reguzzoni2005-be}.

An additional scaled observation noise $I \cdot \alpha \cdot \text{(median variance)}$ helps stabilize iterations of the CL method when the covariance matrix has small or vanishing eigenvalues; this modification was implemented in the batch selection in FolDE where $\alpha$ is set to 6.0. We can interpret this added noise as a minimum observation noise assumed for any candidate and their imagined value, where every imagined value (“lie”) has a base level of uncertainty. A very low base level of uncertainty means that the lie is confident: the CL method will be pessimistic and explore aggressively. A higher base level of uncertainty weakens the lie: the batch selection will balance exploration and exploitation, as the model is not confident about the pessimistic lies. Thus, the scaled observation noise alpha is a knob to tune the degree of exploration and exploitation tradeoff.

\subsection{Benchmarking and Evaluation}

\textbf{Baselines} We compared FolDE against three baselines corresponding to commonly used protein optimization approaches. The first is random selection, typically used in directed evolution of proteins when it is possible to screen thousands or millions of sequences. The second is zero-shot naturalness-based selection, corresponding to picking successive batches of the top ranked naturalness mutants. This represents the scenario where PLMs are used for selection but measurements are not incorporated into future predictions. The third is a random forest model, which corresponds to the EVOLVEpro workflow \citep{Jiang2025-pb}. Jiang et al. systematically evaluated many known few-shot approaches and found that this configuration was the best-performing. We implemented the EVOLVEpro workflow faithfully, specifically: the first round is selected randomly; later rounds train an activity prediction model using a common few-shot architecture (ESM2-15b embeddings + a random forest top layer), and select a batch from the mutants with the top predicted activity. Given their experimental success and thorough testing of existing methods, we regard EVOLVEpro as the forefront workflow for low-N protein optimization.

\textbf{ProteinGym Simulations} We evaluated FolDE using deep mutational scanning data from ProteinGym, splitting datasets into training (9 single-mutation datasets) and test (17 single-mutation and 3 multi-mutation) sets.

The training set of proteins was selected by filtering ProteinGym datasets with the following criteria: (A) is not a virus, (B) has more than 4000 single mutant measurements, (C) is not a multi-mutation dataset, (D) is not of type “Expression” or “Binding”, (E) was not used to train EVOLVEpro. From those, we chose 9 (Sup. Table 1) which seemed difficult to engineer. We estimated engineering difficulty by the relative activity of a mutant in the 99.375th percentile mutant in the dataset, corresponding to 10 rounds of random selection of 16 mutants: $\text{difficulty}=\frac{A_{99.375\%}-A{\text{min}}}{A_{\text{max}}-A_{\text{min}}}$.

The test single-mutation dataset was composed of the other 17 single mutants satisfying requirements A-E.

For the multi-mutation benchmark, we selected datasets which most closely matched the search space encountered in real campaigns. Specifically, we were interested in the model’s ability to disambiguate large numbers of mutations that have mutations in common. For each of the multi-mutation datasets in ProteinGym, we looked at how often each single mutation occurred in multiple mutants. We selected three multi-mutation datasets for which the average number of multi-mutants derived from each single mutation was greater than 50.

Our simulations approximately follow previously described iterative evaluations for protein optimization workflows as in \cite{Hawkins-Hooker2024-il} and \cite{Notin2023-vj}. We make three modifications. First, we held out 50\% of mutants in each simulation run, which enables simulation bootstrapping. Second, we do not provide the model with any “initial labeled data” - as of round-1, no data has been collected, since we are including zero-shot acquisition functions in our workflow. Third, we compared methods on the basis of “number of top 10\% mutants discovered” and “probability of finding at top 1\% mutant,” instead of the recall@30\% used in \cite{Notin2023-vj} or recall@100 used in \cite{Hawkins-Hooker2024-il}.

\textbf{Benchmarks} Simulations ran for 3 rounds of 16 mutants each, evaluating performance using cumulative number of top 10\% mutants and probability of finding a top 1\% mutant as metrics.

\textbf{Simulation Metrics} We report two primary metrics: (1) cumulative top 10\% hits - the total number of top 10\% performing mutants discovered across the first three rounds, and (2) probability of finding a top 1\% mutant - whether at least one top 1\% mutant was discovered within three rounds. These metrics directly measure the success of protein optimization campaigns: finding high-performing mutants efficiently.

\subsection{Statistical Analysis}

For comparing workflows, we determine whether FolDE discovers significantly more top-performing mutants than baseline methods. Specifically, when comparing two workflows A and B, we simulate both workflows on a set of proteins $p \in P$ and measure $N_{\text{top 10\%},p,A}$ and $N_{\text{top 10\%}, p, B}$, the cumulative number of top 10\% mutants discovered in the first three rounds for each protein. We perform a one-sided Wilcoxon signed-rank test on the log-transformed differences  $\text{log}(N_{\text{top 10\%},p,A}) - \text{log}(N_{\text{top 10\%}, p, B})$ to test whether workflow A discovers more top performers than workflow B. Log-transformation gives us the distribution of relative changes in the number of mutants.

Both the single-mutation and multi-mutation benchmarks consist of protein targets that were not used during model building or in the construction of EVOLVEpro.

\section{Ethics Statement}

We evaluated the merits of openly releasing FolDE using the Responsible AI x Biodesign framework. FolDE provides an incremental improvement in protein optimization efficiency without introducing categorically new capabilities. The main beneficiaries are resource-limited labs working on therapeutic development, sustainable biomanufacturing, and pandemic preparedness. In contrast, sophisticated actors with malicious intent may possess the resources to use less efficient protein engineering methods, making FolDE's marginal efficiency gains unlikely to significantly change the threat landscape \citep{Gotting2025-yv}.

As precautionary measures, we excluded viral proteins from benchmark development and do not distribute protein language models or weights with our code, preserving existing access controls \citep{Bloomfield2024-go}. We conclude that open release of FolDE advances scientific progress while presenting minimal incremental risk.

\section{Data Availability}
Users at Lawrence Berkeley National Lab can run the FolDE workflow the LBNL Foldy instance at foldy.lbl.gov. All naturalness runs and embeddings for the single- and multi-mutation ProteinGym targets described here can be downloaded from the LBNL Foldy instance via the folde-train tag or folde-test tag. Simulation results are shipped with the Foldy repository at https://github.com/jbei/foldy.

\section{Software Availability}
The FolDE implementation and simulation software are available under a modified BSD license at https://github.com/jbei/foldy. FolDE is incorporated in a web application, Foldy, which is deployable with a single command on personal-computers, but benefits from GPU acceleration. Through the Foldy web app, users can manage protein optimization campaigns, using FolDE or other workflows. 

\bibliographystyle{unsrtnat}
\bibliography{references}  %%% Uncomment this line and comment out the ``thebibliography'' section below to use the external .bib file (using bibtex) .

% \bibliography{}

\section{End Notes}
\subsection{Acknowledgments}
We thank Justin Baerwald for his assistance with automation, Qingyun Dan for his advice on protein optimization, Lucas Waldburger for his advice on machine learning, Nathan Lanclos and Bridge Cawthon for help supporting the Foldy website, and Reginald Roberts for his help with editing the manuscript.

\subsection{Author Contributions}
\textbf{Conceptualization:} J.B.R., C.R.J., I.D., T.D.Y., M.C.T.A., B.E., J.D.K. \textbf{Formal Analysis:} J.B.R. \textbf{Funding Acquisition:} B.E., J.D.K. \textbf{Methodology:} J.B.R., C.R.J., I.D., T.D.Y., B.E. \textbf{Software:} J.B.R., C.R.J., I.D. \textbf{Supervision:} B.E., J.D.K. \textbf{Visualization:} J.B.R. \textbf{Writing - Original Draft:} J.B.R., C.R.J., T.D.Y., A.N.P., B.E., J.D.K. \textbf{Writing - Review \& Editing:} J.B.R., C.R.J., I.D., T.D.Y., A.N.P., G.A.H., L.S.K., S.H.K., P.H.W., J.L., M.W., I.V.S., M.C.T.A., B.E., J.D.K.

\subsection{Funding}
J.B.R. was supported in part by a fellowship award under contract [FA9550-21-F-0003] through the National Defense Science and Engineering Graduate (NDSEG) Fellowship Program, sponsored by the Air Force Research Laboratory (AFRL), the Office of Naval Research (ONR) and the Army Research Office (ARO). C.R.J. was supported in part by a fellowship award under Grant No. DGE2444107 through the National Science Foundation Graduate Research Fellowship Program (NSF GRFP). Any opinions, findings, and conclusions or recommendations expressed in this material are those of the authors and do not necessarily reflect the views of the National Science Foundation.

This work was part of the DOE Joint BioEnergy Institute (https://www.jbei.org) supported by the U.S. Department of Energy, Office of Science, Office of Biological and Environmental Research, supported by the U.S. Department of Energy, Energy Efficiency and Renewable Energy, Bioenergy Technologies Office, through contract DE-AC02-05CH11231 between Lawrence Berkeley National Laboratory and the U.S. Department of Energy.

Research reported in this publication was supported by the National Institute Of General Medical Sciences of the National Institutes of Health under Award Numbers R01AI186111 and F32GM153046. The content is solely the responsibility of the authors and does not necessarily represent the official views of the National Institutes of Health.

The views and opinions of the authors expressed herein do not necessarily state or reflect those of the United States Government or any agency thereof. Neither the United States Government nor any agency thereof, nor any of their employees, makes any warranty, expressed or implied, or assumes any legal liability or responsibility for the accuracy, completeness, or usefulness of any information, apparatus, product, or process disclosed, or represents that its use would not infringe privately owned rights.

\subsection{Competing Interests}
J.B.R has financial interests in AlkaLi Labs. J.D.K. has financial interests in Ansa Biotechnologies, Apertor Pharma, Berkeley Yeast, Demetrix, Lygos, Napigen, ResVita Bio, and Zero Acre Farms.

\clearpage
\begin{supplement}
\section*{Supplementary Information}

\begin{table}[ht]
\centering
% \captionsetup{labelformat=empty} % hide "Table S1:"
\caption{ProteinGym Datasets Used in Study}
\begin{tabularx}{\linewidth}{l l r X}
\hline
\textbf{Set} & \textbf{ProteinGym ID} & \textbf{Number of Mutants} & \textbf{Notes} \\ \hline
Train & ANCSZ\_Hobbs\_2022 & 4670 & \\ \hline
Train & BLAT\_ECOLX\_Firnberg\_2014 & 4783 & \\ \hline
Train & CBS\_HUMAN\_Sun\_2020 & 7217 & \\ \hline
Train & HEM3\_HUMAN\_Loggerenberg\_2023 & 5689 & \\ \hline
Train & HSP82\_YEAST\_Flynn\_2019 & 13294 & \\ \hline
Train & hKKA\_HUMAN\_Gersing\_2022\_activity & 8570 & \\ \hline
Train & OXDA\_RHOTO\_Vanella\_2023\_activity & 6396 & \\ \hline
Train & PPM1D\_HUMAN\_Miller\_2022 & 7889 & \\ \hline
Train & SHOQ2\_HUMAN\_Kwon\_2022 & 10972 & \\ \hline
Single Mutation & ADGRB6\_PSEA1\_Chen\_2020 & 5004 & \\ \hline
Single Mutation & AMIE\_PSEAE\_Wrenbeck\_2017 & 6227 & \\ \hline
Single Mutation & CAS9\_STRP1\_Spencer\_2017\_positive & 8117 & \\ \hline
Single Mutation & HMDH\_HUMAN\_Jiang\_2019 & 16853 & \\ \hline
Single Mutation & KCNJ2\_MOUSE\_Coyote-Maestas\_2022\_function & 6963 & \\ \hline
Single Mutation & KKAA\_KLEPN\_Melnikov\_2014 & 4960 & \\ \hline
Single Mutation & LGK\_LIPST\_Klesmith\_2015 & 7890 & \\ \hline
Single Mutation & MET\_HUMAN\_Eastew\_2023 & 5393 & \\ \hline
Single Mutation & MLAC\_ECOLI\_MacRae\_2023 & 4007 & \\ \hline
Single Mutation & MSH2\_HUMAN\_Jia\_2020 & 16749 & \\ \hline
Single Mutation & MTHR\_HUMAN\_Weile\_2021 & 12464 & \\ \hline
Single Mutation & PAH\_HUMAN\_Huttinger\_2021 & 5345 & \\ \hline
Single Mutation & PPARG\_HUMAN\_Majithia\_2016 & 9576 & \\ \hline
Single Mutation & PTEN\_HUMAN\_Mighell\_2018 & 7260 & \\ \hline
Single Mutation & RNC\_ECOLI\_Weeks\_2023 & 4277 & \\ \hline
Single Mutation & S22A1\_HUMAN\_Yee\_2023\_activity & 1094 & \\ \hline
Single Mutation & SC6A4\_HUMAN\_Young\_2021 & 11576 & \\ \hline
Multiple Mutation & GRB2\_HUMAN\_Faure\_2021 & 63366 & 120 double mutants per single mutant \\ \hline
Multiple Mutation & PABP\_YEAST\_Melamed\_2013 & 37708 & 61 double mutants per single mutant \\ \hline
Multiple Mutation & SPG1\_STRSG\_Olson\_2014 & 536962 & 1026 double mutants per single mutant \\ \hline
\end{tabularx}
\end{table}

\begin{figure}
	\centering
	% \fbox{\rule[-.5cm]{4cm}{4cm} \rule[-.5cm]{4cm}{0cm}}
    \includegraphics{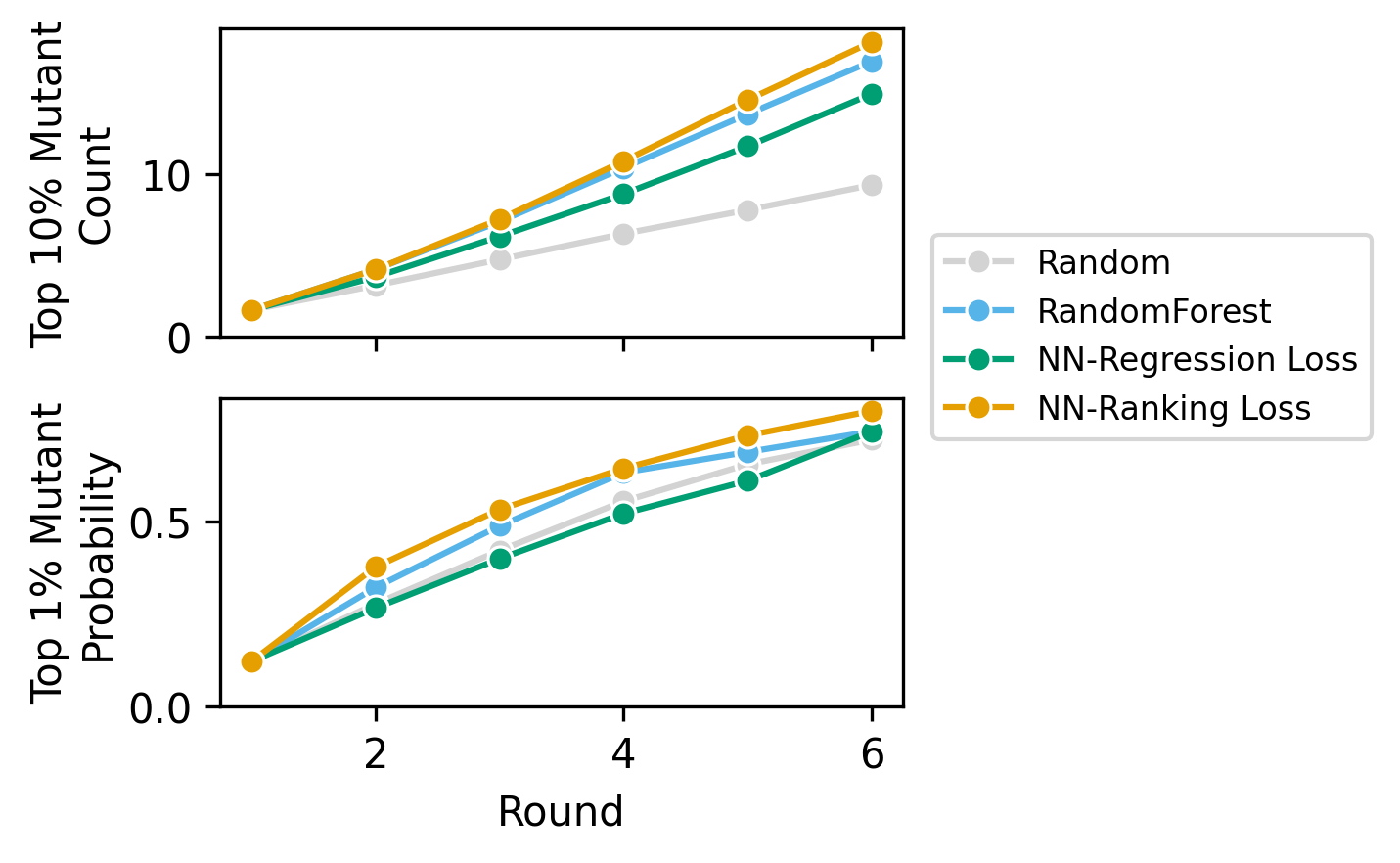}
	\caption{\textbf{Top Layer Architectures} Prediction quality of three top-layer architectures: random forest, a neural network trained with mean squared error loss, and a neural network trained with ranking loss. Evaluated on the training benchmark. All have random mutants selected in round-1.
}
	\label{fig:supfig1}
\end{figure}

\begin{figure}
	\centering
	% \fbox{\rule[-.5cm]{4cm}{4cm} \rule[-.5cm]{4cm}{0cm}}
    \includegraphics{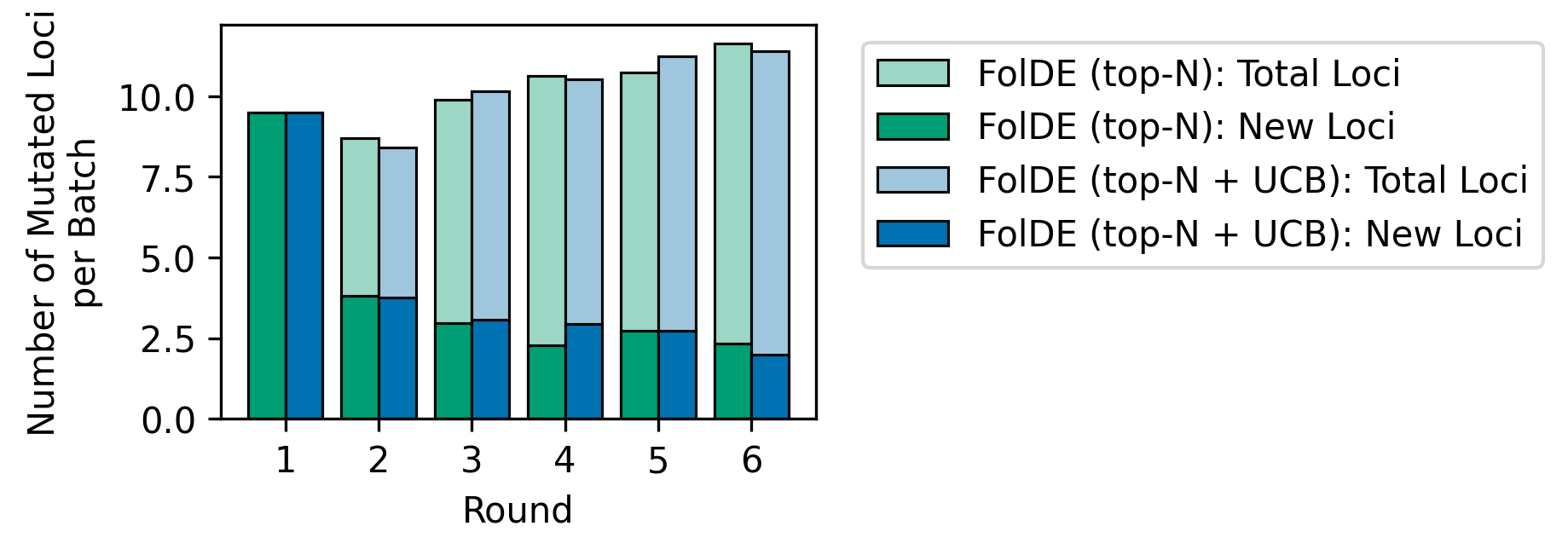}
	\caption{\textbf{Upper Confidence Bound (UCB) Does Not Improve Batch Diversity} The number of mutated loci per batch on the multi-mutation benchmark decreases over rounds for both standard top-N selection and top-N with UCB. Dark bars show new loci not previously mutated; light bars show previously mutated loci. Both selection methods show identical patterns of declining exploration, with the majority of later-round mutations concentrated on previously successful loci. Data averaged across multi-mutation benchmark proteins.
}
	\label{fig:supfig2}
\end{figure}

\begin{figure}
	\centering
	% \fbox{\rule[-.5cm]{4cm}{4cm} \rule[-.5cm]{4cm}{0cm}}
    \includegraphics[]{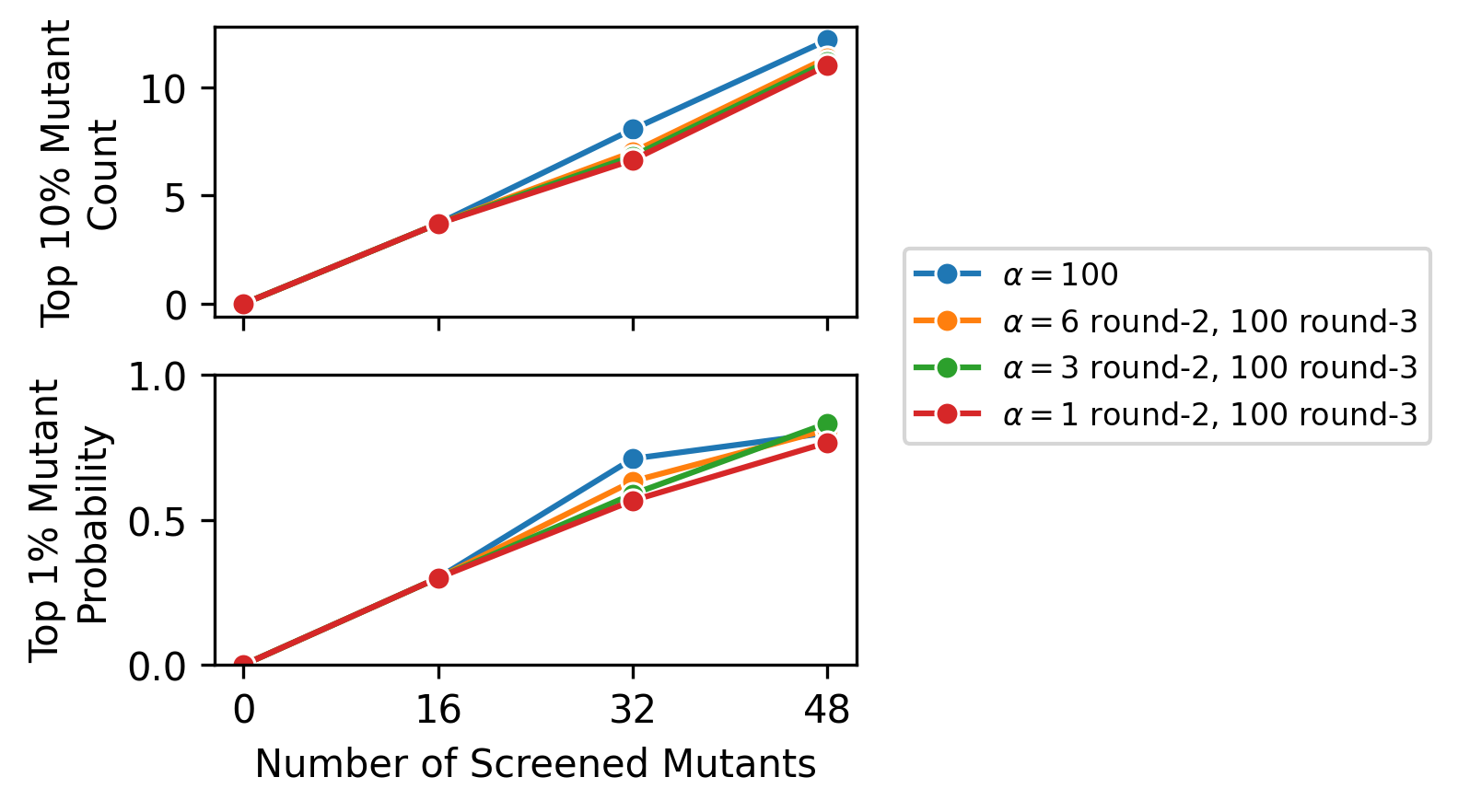}
	\caption{\textbf{Optimizing Constant-Liar $\alpha$ For Three Round Campaign} More aggressive constant-liar (lower $\alpha$) worsens the number of top 10\% mutants discovered and, for some values, improves the 3-round probability of finding a top 1\% mutant.
}
	\label{fig:supfig3}
\end{figure}

\begin{figure}
	\centering
	% \fbox{\rule[-.5cm]{4cm}{4cm} \rule[-.5cm]{4cm}{0cm}}
    \includegraphics[width=\linewidth]{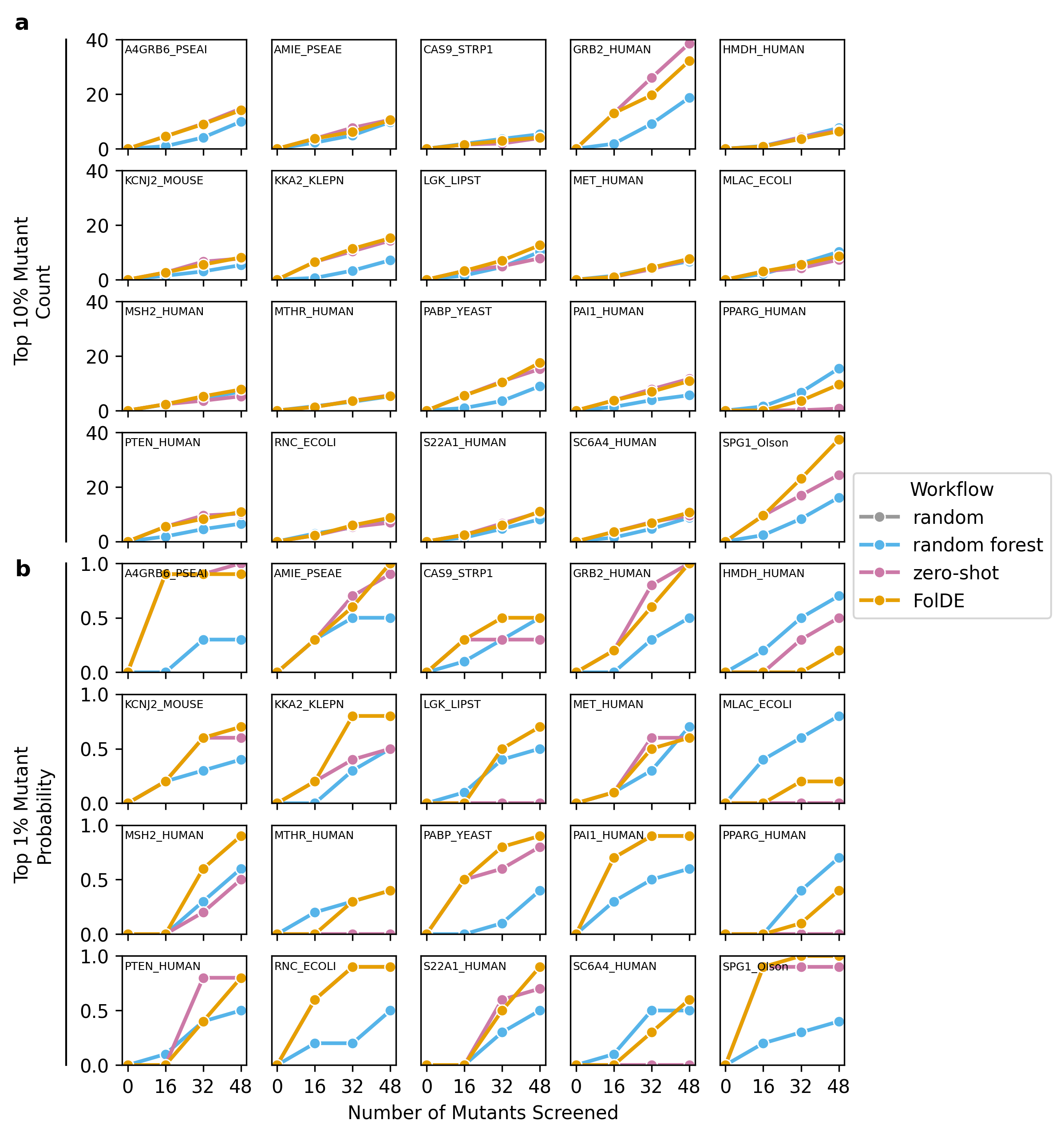}
	\caption{\textbf{FolDE Comparison to Baseline, Per-Target, for a Three Round Campaign} (a) the number of top 10\% performing mutants discovered and (b) probability of finding a top 1\% mutant.
}
	\label{fig:supfig4}
\end{figure}

\begin{figure}
	\centering
	% \fbox{\rule[-.5cm]{4cm}{4cm} \rule[-.5cm]{4cm}{0cm}}
    \includegraphics{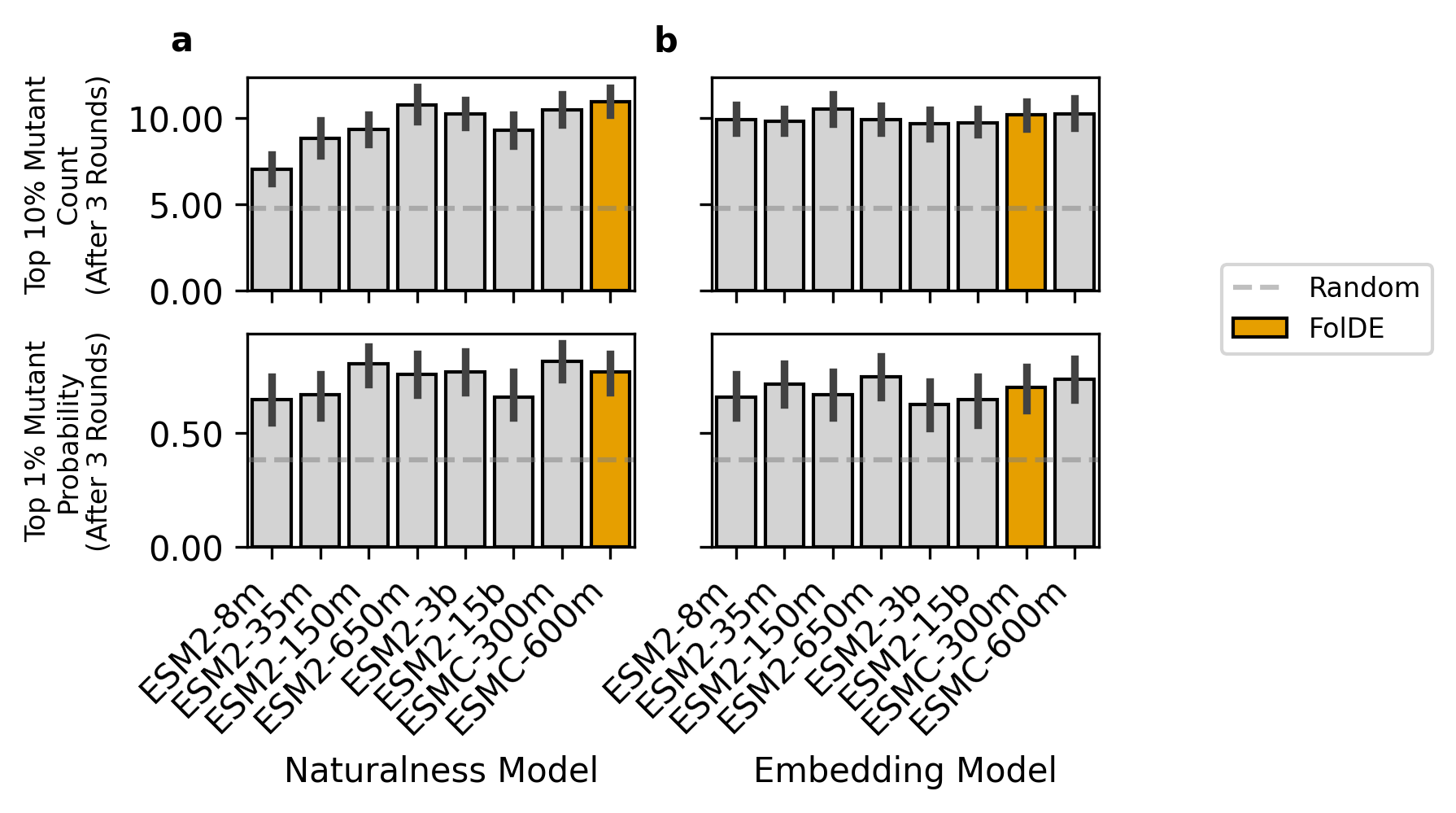}
	\caption{\textbf{Protein Language Model Sizing} (a) Prediction quality of the PLM as a zero-shot activity predictor was evaluated on the training set of proteins, as measured by both the cumulative number of top 10\% mutants discovered after 3 rounds (top) and the probability of finding a top 1\% mutant (bottom). (b) Similarly, the performance of the FolDE model was evaluated on round-3 for different PLM embedding models.
}
	\label{fig:supfig5}
\end{figure}

\begin{figure}
	\centering
	% \fbox{\rule[-.5cm]{4cm}{4cm} \rule[-.5cm]{4cm}{0cm}}
    \includegraphics{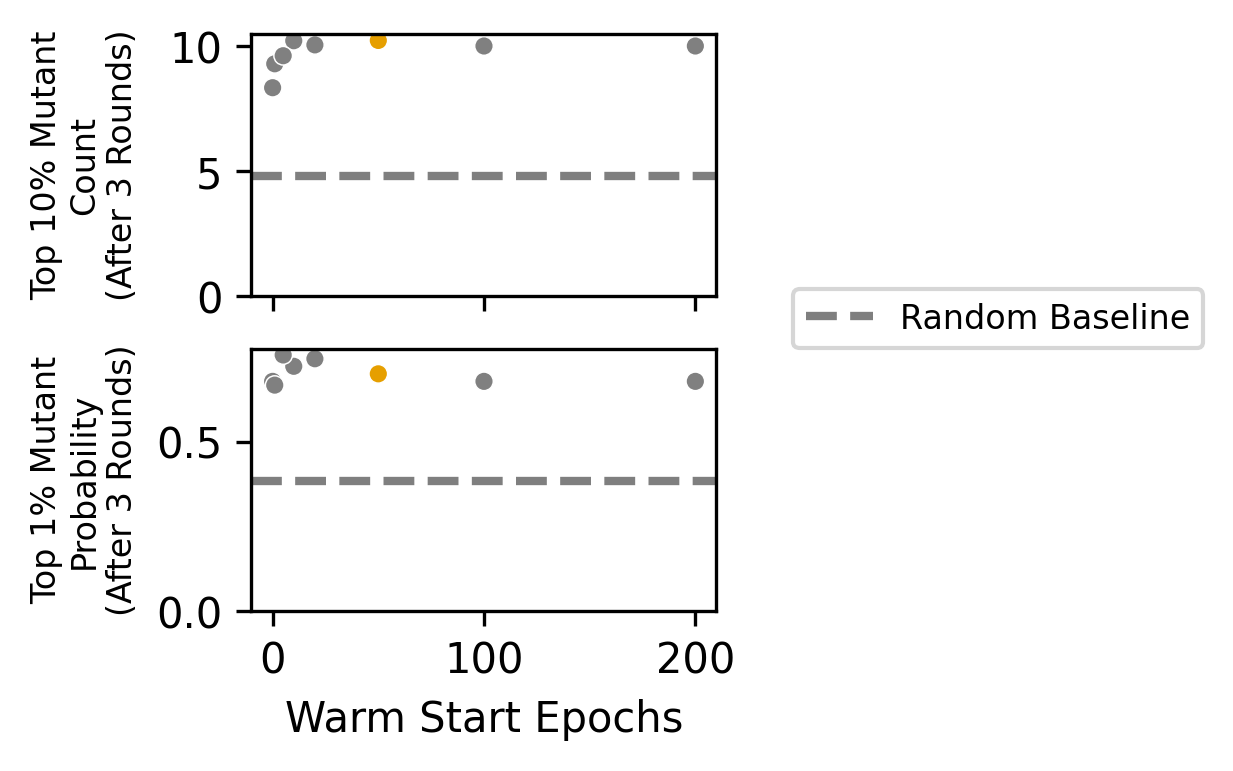}
	\caption{\textbf{Calibrating the Naturalness Warm-Start} For round-3 evaluated on the training dataset, a sweep over warm-start training epochs shows that the improvements have stabilized after about 10 epochs, as measured by both cumulative top 10\% mutants and probability of finding a top 1\% mutant. Evaluated on 0, 1, 5, 10, 20, 50, 100, 200 epochs.
}
	\label{fig:supfig6}
\end{figure}

\end{supplement}

\end{document}